\documentclass[11pt,xcolor={dvipsnames}]{article}

\usepackage[round]{natbib} 

\usepackage[final]{acl}

\usepackage{times}
\usepackage{latexsym}

\usepackage[T1]{fontenc}

\usepackage[utf8]{inputenc}

\usepackage{microtype}

\usepackage{inconsolata}

\usepackage{graphicx}
\graphicspath{ {./images/} }

\usepackage{booktabs}
\usepackage{multirow}
\usepackage{colortbl}

\usepackage{changepage}

\usepackage[toc,page]{appendix}

\usepackage{rotating}

\usepackage{array} 
\usepackage{ragged2e}
\usepackage{pbox}

\usepackage{listings}
\usepackage{arydshln}

\newcolumntype{L}[1]{>{\RaggedRight\arraybackslash}p{#1}}

\newcommand*{\MinNumber}{0.0}%
\newcommand*{\MidNumber}{0.5}%
\usepackage{tikz}
\usepackage{collcell}
\usepackage{xstring}

\newcommand{\ApplyGradient}[1]{%
    \IfDecimal{#1}{
        \ifdim #1 pt > \MidNumber pt
            \hspace{-1em}{#1}
        \else
            \pgfmathsetmacro{\PercentColor}{max(min(100.0*(\MidNumber - #1)/(\MidNumber-\MinNumber),100.0),0.00)} %
             \hspace{-1em}\textcolor{red}{#1}
        \fi
    }{#1}
}

\newcolumntype{R}{>{\collectcell\ApplyGradient}c<{\endcollectcell}}

\title{SGHateCheck: Functional Tests for Detecting Hate Speech in Low-Resource Languages of Singapore}

\author{Ri Chi Ng, Nirmalendu Prakash, Ming Shan Hee, \\ 
        {\bf Kenny Tsu Wei Choo} \and {\bf Roy Ka-Wei Lee} \\ 
        Singapore University of Technology and Design \\
        richi\_ng@sutd.edu.sg, nirmalendu\_prakash@mymail.sutd.edu.sg, \\ mingshan\_hee@mymail.sutd.edu.sg, kenny\_choo@sutd.edu.sg, roy\_lee@sutd.edu.sg
        }

\begin{document}
\maketitle
\begin{abstract}

To address the limitations of current hate speech detection models, we introduce \textsf{SGHateCheck}, a novel framework designed for the linguistic and cultural context of Singapore and Southeast Asia. It extends the functional testing approach of HateCheck and MHC, employing large language models for translation and paraphrasing into Singapore's main languages, and refining these with native annotators. \textsf{SGHateCheck} reveals critical flaws in state-of-the-art models, highlighting their inadequacy in sensitive content moderation. This work aims to foster the development of more effective hate speech detection tools for diverse linguistic environments, particularly for Singapore and Southeast Asia contexts.

\end{abstract}

{\color{red} \textbf{Disclaimer}: \textit{
This paper contains violent and discriminatory content that may be disturbing to some readers.}} 

\section{Introduction}
Hate speech (HS) detection models have become crucial tools in moderating online content and understanding the dynamics of online hate. Traditionally, these models are evaluated against held-out test sets. However, this method often falls short in fully assessing the models' performance due to the systematic gaps and biases inherent in HS datasets. Recognizing this limitation, functional tests, such as those introduced by HateCheck \cite{rottger-etal-2021-hatecheck} and extended by Multilingual HateCheck (MHC) \cite{rottger-etal-2022-multilingual}, offer a nuanced approach to evaluate HS detection models more thoroughly by simulating a variety of real-world scenarios across multiple languages.

Despite these advancements, there remains a significant gap in HS detection for the diverse linguistic landscape of Singapore. This country is home to a unique mix of commonly used languages, including English, Mandarin Chinese (Mandarin), Tamil, and Malay, each with its own cultural nuances and idiomatic expressions that standard datasets may not fully capture. Furthermore, the Southeast Asian (SEA) cultural context presents additional challenges, as existing models primarily focus on Western cultural contexts, leaving a gap in our understanding and detection capabilities of HS within this region. 

To address these gaps, we introduce \textsf{SGHateCheck}\footnote{Dataset available at https://github.com/Social-AI-Studio/SGHateCheck}, an extension of the HateCheck and MHC frameworks. \textsf{SGHateCheck} is designed to evaluate HS detection models against a comprehensive set of functional tests tailored to the linguistic and cultural nuances of Singapore and the broader SEA context. Through \textsf{SGHateCheck}, we aim to contribute to the development of more inclusive and effective HS detection models, providing better protection against online hate for users in Singapore and SEA. To our knowledge, \textsf{SGHateCheck} is the first functional test comprehensively evaluate HS in Singapore and SEA context.

Similar to MHC, SGHateCheck's functional tests for each language closely align with the original HateCheck's framework, which was developed through interviews with civil society stakeholders and a thorough review of HS research. Unlike MHC, which relied on annotators for manual translation and rewriting of English test cases into other languages, SGHateCheck employs large language models (LLMs) for translating and paraphrasing HateCheck's templates into Singapore's four primary languages. Native language annotators then refine these machine-generated templates. To ensure cultural relevance, we collaborate with experts familiar with Singapore's societal issues to identify vulnerable groups targeted by HS. This information guides the automated generation of test cases, which are further refined by native annotators for accuracy and cultural sensitivity. 

We showcase \textsf{SGHateCheck}'s efficacy as a diagnostic tool by evaluating cutting-edge, fine-tuned LLMs using a mix of publicly available HS datasets in English, Mandarin, and Malay. Although these models perform well on existing datasets, \textsf{SGHateCheck} testing highlights critical limitations: 1) weaker models predominantly misclassified test cases as non-hateful; 2) while multilingual dataset fine-tuning somewhat mitigates biases, the performance gains are modest; 3) compared to MHC and HateCheck, these LLMs underperform on \textsf{SGHateCheck} test cases on selected functionalities, even in languages like English and Mandarin. Such shortcomings could have severe implications if these LLMs were deployed for content moderation, risking harm by inadequately protecting users against HS or unnecessarily limiting free expression. By exposing these flaws, \textsf{SGHateCheck} aims to drive the development of more robust HS detection models, particularly for Singapore and SEA contexts

\section{Constructing \textsf{SGHateCheck}}
\subsection{Defining Hate Speech}

\textsf{SGHateCheck} adopts the definition of HS from HateCheck as ``\textit{abuse aimed at a protected group or its members for belonging to that group.}'' In the context of \textsf{SGHateCheck}, protected groups include race, religion, descent, sexual orientation, gender, disability, and age, aligning with Singapore's legal framework against harmful speech, specifically referencing the Maintenance of Religious Harmony Act\footnote{https://sso.agc.gov.sg/Act/MRHA1990} and the Penal Code's section 298A\footnote{https://sso.agc.gov.sg/Act/PC1871}. These laws serve as the foundation for identifying protected groups in SGHateCheck's functional tests.

\subsection{Defining Functional Tests}

In \textsf{SGHateCheck}, functional tests meticulously evaluate HS detection LLMs by differentiating between various types of hateful and benign content. These tests are designed to be fine-grained, distinct, and contrastive, distinguishing between HS that uses profanity and non-hateful expressions incorporating similar language. For instance, there are specific tests for HS involving profanity against protected groups (e.g., Singlish (SS)\footnote{Singlish refers to the colloquial form of English in Singapore}:``\textit{f*ck all gays lah}'') contrasted with non-hateful, colloquial uses of profanity (e.g.. SS:``\textit{Why these f*cking sirens like that one?}''). Tests are organised into categories based on their thematic relevance, enhancing the models' ability to discern nuanced language use across different contexts.

\subsection{Selecting Functional Tests}

We tailored the selection of functional tests to align closely with the original HateCheck framework, emphasizing engagement with experts on Singapore's societal nuances. Drawing from HateCheck's methodology, which was grounded in interviews with NGO workers and a comprehensive review of HS research, we incorporate Singapore-specific elements. 
This approach enhances the relevance of our tests, making them a robust tool for evaluating HS detection LLMs within Singapore's unique context. All test-cases are short text statements, and they are constructed to be clearly hateful or non-hateful according to our definition of HS.

\textsf{SGHateCheck} comprises 28 functional tests for Singlish, 26 for Mandarin, and 21 each for Malay and Tamil. This customization reflects linguistic and cultural considerations, such as excluding slur homonyms and reclaimed slurs absent in these languages, and omitting spelling variations in Malay and Tamil to simplify translation. For Mandarin, we utilized templates from the Mandarin version of MHC. Like HateCheck and MHC, these tests distinguish between HS and non-hateful content with similar lexical features but clear non-hateful intent, ensuring nuanced detection across diverse expressions of hate.

\textbf{Distinct Expressions of Hate}. \textsf{SGHateCheck} evaluates various forms of HS, including derogatory remarks (\textbf{F1-4}) and threats (\textbf{F5/6}), alongside hate conveyed through slurs (\textbf{F7}) and profanity (\textbf{F8}). It assesses hate articulated via pronoun references (\textbf{F10/11}), negation (\textbf{F12}), and different phrasings like questions and opinions (\textbf{F14/15}). Uniquely, it includes tests for Singlish, featuring spelling variations such as omissions or leet speak (\textbf{F23-34}), and for Mandarin, it considers non-Latin script variations and Pinyin spelling (\textbf{F32-34}), enriching its evaluative scope.


\textbf{Contrastive Non-Hate}. \textsf{SGHateCheck} also evaluates non-hateful content, including uses of profanity (\textbf{F9}), negation (\textbf{F13}), and references to protected groups without malice (\textbf{F16/17}). It further examines contexts where HS is quoted or countered, specifically in counter-speech scenarios where responses aim to neutralize hate (\textbf{F18/19}). Additionally, it differentiates content targeting non-protected entities, such as objects (\textbf{F20-22}), ensuring a clear distinction between HS and non-hateful.

\subsection{Generating Test-Cases}

We adapted HateCheck's test cases for Singlish, Malay, and Tamil using a combination of machine translations from ChatGPT and Google Translate, followed by rigorous review and adjustment by bilingual translators. Initially, we applied these translation tools to adapt HateCheck templates for the mentioned languages, while Mandarin test cases were directly sourced from MHC. Subsequently, bilingual translators were enlisted to validate and refine these translations, including the MHC Mandarin templates, ensuring accuracy and cultural relevance. This process involved at least two translators per language, allowing for a nuanced approach to translation, particularly for terms without direct equivalents, such as certain profanities in Malay and Tamil. Table~\ref{tab:translation_metrics} summarizes the distribution of machine-translated templates verified and accepted by the translators, as well as the number of manually rewritten templates. In these test cases, templates translated by ChatGPT accounted for approximately half of all templates utilized. This suggests significant potential for leveraging machine translations to assist in the translation of templates across other low-resource languages


Templates that were not machine-translated were manually created by the translators. Importantly, translators were not required to adhere rigidly to the English originals. For instance, templates containing the word 'f*ck'—which lacks a direct Malay equivalent—were either adapted to an alternative expression ('Kau boleh pergi mampus selamanya', translating to 'die forever' in English, for 'You should f*ck off forever'), or they were excluded from the list of templates, as indicated by 'Dropped' in Table~\ref{tab:translation_metrics}. All templates inherit the predetermined hatefulness from HateCheck and MHC.


Next, we leverage the constructed good quality templates to generate the test cases for the protected groups. In \textsf{SGHateCheck}, protected groups are delineated by specific target groups (e.g., Hindus within the category of Religion), using placeholders in test cases to denote group identifiers. Templates facilitate the scalable generation of test cases, seamlessly replacing placeholders for both group identifiers (e.g.. SS:``\textit{\{IDENTITY\_P\} are a pest.}'') and common slurs (e.g.. SS:``\textit{The \{SLUR\_P\} sia, always like that one.}''), streamlining the customization process for each test case. Target groups and common slurs were identified with consultation of local experts. This method ensures comprehensive representation across all protected groups without the need for additional, language-specific placeholders. Each generated test case is verified by two to three native speakers who are proficient in the languages for validation later on. \textsf{SGHateCheck} uniformly covers six protected groups across all languages, reflecting consistent social contexts and targets, thereby maintaining uniformity in addressing HS across diverse linguistic settings.

In total, across four languages, \textsf{SGHateCheck} comprises 21,152 test cases, with 15,052 classified as hateful and 6,100 as non-hateful according to the template labels. The distribution varies by language due to differing numbers of functional tests and slurs, with Singlish featuring the highest number of cases (7,023) and Tamil the fewest (2,851). The average length of a test case is 10.5 words or 42.6 characters, showcasing the dataset's diversity and depth.

\begin{table}[t]
\small
\centering
\begin{tabular}{rccc}
\toprule
                              & Singlish    & Malay       & Tamil       \\
\midrule
ChatGPT & 371 & 358 & 193 \\
Google Transl. & - & 96 & 61 \\
HateCheck & 77 & - & - \\
Manual Written   & 153 & 209 & 227 \\
Dropped  & 0 & 8 & 12 \\
\midrule
Total & 601 & 671 & 399 \\
\bottomrule
\end{tabular}
\caption{Distribution of template translation for Singlish, Malay and Tamil}
\label{tab:translation_metrics}
\end{table}

\subsection{Validation}

Each test case is associated with a predefined gold label from its corresponding template, indicating its level of hatefulness. A total of 10,926 test cases were sampled and annotated by 16 recruited annotators to ensure the quality and accuracy of the data. Each test case was reviewed by three annotators for English, Malay, and Mandarin languages and by two annotators for Tamil language.
Annotators followed specific guidelines to maintain a consistent definition of hate. To ensure that only high quality test cases were used in the experiments, test cases lacking majority agreement or mismatching their gold label were excluded from further experiments. Consequently, 10,394 test cases were retained for the study, while 532 were excluded.
The inter-annotator agreement and excluded test cases can be found in Appendix \ref{appendix:iaa}.



\section{Benchmarking LLMs on \textsf{SGHateCheck}}
We evaluated various state-of-the-art open-source LLMs such as mBERT, LLaMA2, SEA-LION, and SeaLLM using SGHateCheck. These LLMs were fine-tuned with existing hate speech datasets before testing.

The BERT multilingual base model (uncased) (mBERT)~\cite{mbert} employs masked language modeling (MLM) and next sentence prediction (NSP) for its training. It supports 104 languages, prominently including English, Mandarin, Tamil, and Malay, facilitating a broad linguistic reach for applications in diverse linguistic environments.

The LLaMA2 model~\cite{llama_2}, part of Meta's auto-regressive LLM family, is available in sizes ranging from 7 to 70 billion parameters. We utilized the 7 billion parameter version. Predominantly trained on English (89.7\%), it includes minor language data contributions (0.01-0.17\%).

The Mistral-7B model~\cite{mistral} is an auto-regressive model noted for its performance, outpacing LLaMA in tasks like content moderation. Although the specifics of its training data are not disclosed, it has shown effectiveness in Southeast Asian languages.

The SEA-LION-7B model~\cite{sealion}, leveraging the MPT architecture, is specifically trained on a wide array of languages from the Southeast Asian region, including Thai, Vietnamese, Indonesian, Chinese, Khmer, Lao, Malay, Burmese, Tamil, and Filipino, showcasing its focus on linguistic diversity within this geographic area.

The SeaLLMv1-7B model~\cite{seallm}, developed on the LLaMA2 architecture, underwent initial pre-training with a dataset comprising English and several Southeast Asian languages, including Thai, Vietnamese, Indonesian, Chinese, Khmer, Lao, Malay, Burmese, and Tagalog. It was then fine-tuned with a similar language set, albeit with an increased emphasis on English content, to enhance its linguistic versatility and performance.

\subsection{LLM Fine-tuning}

We devised two specialized datasets, EngSet and MultiSet, tailored for training the benchmark LLMs to recognize HS across different linguistic contexts. EngSet integrates English-language data from two prominent sources, Twitter Hate~\cite{waseem-hovy-2016-hateful} and HateXplain~\cite{mathew2021hatexplain}, to capture a wide range of hateful and non-hateful content. MultiSet expands this framework into a multilingual domain by incorporating Mandarin and Malay examples from COLD~\cite{chinese_training_dataset} and HateM~\cite{malay_dataset}, respectively, creating a richer dataset that reflects the linguistic diversity encountered in HS detection. For each of these sets, we use part of the data for fine-tuning and a held out set for evaluation. We use the binary (hateful or non-hateful) labels to fine-tune the LLMs using LoRA~\cite{lora} adapter training except for mBERT, which we perform full fine-tuning.

To assess the efficacy of the LLMs, held-out tests were conducted using samples from COLD (in MultiSet) and HateXplain (in both EngSet and MultiSet). The results, detailed in Table \ref{tab:held_out}, indicate that most LLMs achieved commendable performance, with accuracy and F1 scores ranging from 0.7 to 0.9. SEA-LION was the outlier, with its scores falling below the 0.7 threshold across all evaluated metrics, highlighting a potential area for improvement in handling diverse linguistic data.


\begin{table*}[]
\small
\centering
\def\width{15mm}
\begin{tabular}{>{\arraybackslash}m{1.5cm}>{\arraybackslash}m{2cm}*{5}{Rc}}
\toprule
       \multirow{2}{1.5cm}{Fine-tune Dataset}&  \multirow{2}{2cm}{Held-out Dataset}   & \multicolumn{2}{c}{LL} & \multicolumn{2}{c}{MB} & \multicolumn{2}{c}{MI} & \multicolumn{2}{c}{SO} & \multicolumn{2}{c}{SM} \\
    \cmidrule(lr){3-4} \cmidrule(lr){5-6} \cmidrule(lr){7-8} \cmidrule(lr){9-10} \cmidrule(lr){11-12} 
& & Acc. & F1 & Acc. & F1 & Acc. & F1 & Acc. & F1 & Acc. & F1 \\
\midrule
EngSet & HateExplain & 0.835  & 0.723 & 0.837 & 0.745 & 0.851 & 0.756 & 0.667 & 0.065 & 0.836 & 0.725   \\
MultiSet & HateExplain & 0.834 & 0.728 & 0.845 & 0.753 & 0.831 & 0.704 & 0.685 & 0.192 & 0.802 & 0.657  \\
MultiSet & COLD       & 0.797 & 0.719 & 0.809 & 0.781 & 0.796 & 0.763 & 0.533 & 0.378 & 0.783 & 0.749  \\
\bottomrule
\end{tabular}
\caption{Accuracy (Acc.) and F1 for held-out tests, for LL:LLaMA2, MB:mBert, MI:Mistral, SO:SEA-LION and SM: SeaLLM. }
\label{tab:held_out}
\end{table*}


\begin{table*}[]
    \small
    \centering
    {
    \begin{tabular}{>{\arraybackslash}m{1.5cm}>{\arraybackslash}m{1.5cm}*{10}{R}}
       \toprule
       
       \multirow{2}{1.5cm}{Metric}&
       \multirow{2}{1.5cm}{Fine-tune Dataset}
        & \multicolumn{2}{c}{Average}
        & \multicolumn{2}{c}{Singlish}
        & \multicolumn{2}{c}{Malay}
        & \multicolumn{2}{c}{Mandarin}
        & \multicolumn{2}{c}{Tamil}
        \\
        \cmidrule(lr){3-4} \cmidrule(lr){5-6}\cmidrule(lr){7-8} \cmidrule(lr){9-10} \cmidrule(lr){11-12} &
        & {NH} & {H}
        & {NH} & {H}
        & {NH} & {H}
        & {NH} & {H}
        & {NH} & {H} \\

        \midrule
        \multirow{2}{1.5cm}{Accuracy}&
        EngSet & 0.981 & 0.108 & 0.952 & 0.277 & 0.991
        & 0.087 & 0.996 & 0.060 & 0.986 & 0.008
        \\
        &MultiSet & 0.784 & 0.413 & 0.842 & 0.455 & 0.705
        & 0.502 & 0.624 & 0.636 & 0.965 & 0.058 
        \\
        \midrule
        \multirow{2}{1.5cm}{F1} &
        EngSet & \multicolumn{2}{c}{0.307} & \multicolumn{2}{c}{0.404} & \multicolumn{2}{c}{0.309}
        & \multicolumn{2}{c}{0.263} & \multicolumn{2}{c}{0.252}
        \\
        &MultiSet & \multicolumn{2}{c}{0.480} & \multicolumn{2}{c}{0.507} & \multicolumn{2}{c}{0.536}
        & \multicolumn{2}{c}{0.585} & \multicolumn{2}{c}{0.291}
        \\
        \bottomrule
    \end{tabular}
    }
    \caption{Average accuracy and F1 for cases labeled non-hateful (NH) and hateful (H) for each language averaged across the fine-tuned LLMs. Red numbers indicate an accuracy of less than 0.500, which is worse than chance.}
    \label{tab:label_results}
\end{table*}

\subsection{How do the models perform overall?}
Table~\ref{tab:label_results} shows the average accuracy and F1 scores across the benchmark LLMs. SGHateCheck's analysis illustrates a performance discrepancy between LLMs fine-tuned on EngSet, a monolingual dataset, and those on MultiSet, a multilingual dataset. EngSet-tuned models, with a significantly lower average macro F1 score, predominantly misclassify test cases as non-hateful, resulting in a skewed accuracy favoring non-hateful classifications. This imbalance highlights the models' limitations in effectively detecting HS within monolingual data, underscoring the enhanced performance and adaptability of LLMs fine-tuned on multilingual datasets.  Conversely, MultiSet-tuned models show more balanced accuracy across languages but vary in performance by language, with Tamil displaying notably low F1 scores attributed to a high bias. The LLMs achieve the highest F1 scores for Mandarin tests, suggesting better model generalization for this language.

\subsection{How do the fine-tuned models perform across Functional Tests?}


\begin{sidewaystable*}
\small
\begin{tabular}{ll*{4}{|*{5}{R}}}
\toprule
                       &             & \multicolumn{5}{c}{Singlish Accuracy}        & \multicolumn{5}{c}{Malay Accuracy}           & \multicolumn{5}{c}{Mandarin Accuracy}        & \multicolumn{5}{c}{Tamil Accuracy}            \\ \cmidrule(lr){3-7} \cmidrule(lr){8-12} \cmidrule(lr){13-17} \cmidrule(lr){18-22}
F\#                    & GL  & LL & MB & MI & SO & SM & LL & MB & MI & SO & SM & LL & MB & MI & SO & SM & LL & MB & MI & SO & SM \\
\hline
F1   & H & 0.326 & 0.246 & 0.681 & 0.014 & 0.601 & 0.608 & 0.240 & 0.576 & 0.120 & 0.872 & 0.892 & 0.446 & 0.993 & 0.058 & 0.928 & 0.000 & 0.023 & 0.038 & 0.000 & 0.000 \\
F2   & H & 0.408 & 0.112 & 0.847 & 0.000 & 0.765 & 0.620 & 0.463 & 0.639 & 0.046 & 0.954 & 0.794 & 0.431 & 1.000 & 0.157 & 0.971 & 0.000 & 0.057 & 0.109 & 0.000 & 0.000 \\
F3   & H & 0.455 & 0.186 & 0.800 & 0.000 & 0.766 & 0.508 & 0.308 & 0.523 & 0.015 & 0.954 & 0.875 & 0.352 & 0.984 & 0.070 & 0.922 & 0.000 & 0.058 & 0.130 & 0.000 & 0.000 \\
F4   & H & 0.459 & 0.189 & 0.711 & 0.006 & 0.516 & 0.507 & 0.271 & 0.243 & 0.100 & 0.650 & 0.566 & 0.184 & 0.846 & 0.279 & 0.640 & 0.000 & 0.056 & 0.176 & 0.000 & 0.040 \\
F5   & H & 0.519 & 0.130 & 0.878 & 0.000 & 0.702 & 0.470 & 0.289 & 0.417 & 0.104 & 0.922 & 0.755 & 0.230 & 0.942 & 0.137 & 0.921 & 0.000 & 0.057 & 0.100 & 0.000 & 0.000 \\
F6   & H & 0.703 & 0.464 & 0.964 & 0.007 & 0.935 & 0.647 & 0.489 & 0.619 & 0.079 & 0.978 & 0.829 & 0.464 & 0.950 & 0.214 & 0.929 & 0.000 & 0.134 & 0.244 & 0.000 & 0.000 \\
F7   & H & 0.600 & 0.000 & 0.500 & 0.000 & 0.500 & 0.556 & 0.556 & 0.333 & 0.000 & 0.778 & 1.000 & 0.400 & 0.800 & 0.000 & 0.600 & 0.000 & 0.125 & 0.000 & 0.000 & 0.000 \\
F8   & H & 0.579 & 0.214 & 0.793 & 0.007 & 0.779 & 0.629 & 0.500 & 0.536 & 0.064 & 0.886 & 0.928 & 0.667 & 1.000 & 0.232 & 0.877 & 0.000 & 0.045 & 0.099 & 0.000 & 0.000 \\
F9   & N & 0.889 & 1.000 & 1.000 & 1.000 & 1.000 & 0.400 & 0.700 & 1.000 & 1.000 & 0.800 & 0.500 & 0.700 & 0.800 & 0.800 & 0.800 & 1.000 & 1.000 & 1.000 & 1.000 & 1.000 \\
F10  & H & 0.561 & 0.216 & 0.827 & 0.079 & 0.791 & 0.529 & 0.336 & 0.471 & 0.121 & 0.914 & 0.882 & 0.529 & 1.000 & 0.206 & 0.860 & 0.000 & 0.029 & 0.279 & 0.000 & 0.019 \\
F11  & H & 0.529 & 0.436 & 0.821 & 0.043 & 0.793 & 0.657 & 0.686 & 0.757 & 0.064 & 0.957 & 0.914 & 0.396 & 1.000 & 0.295 & 0.986 & 0.000 & 0.256 & 0.467 & 0.000 & 0.006 \\
F12  & H & 0.414 & 0.081 & 0.901 & 0.018 & 0.784 & 0.421 & 0.325 & 0.526 & 0.070 & 0.807 & 0.799 & 0.302 & 0.986 & 0.122 & 0.906 & 0.000 & 0.088 & 0.252 & 0.000 & 0.000 \\
F13  & N & 0.871 & 0.823 & 0.895 & 0.992 & 0.887 & 0.608 & 0.672 & 0.856 & 0.904 & 0.688 & 0.555 & 0.609 & 0.609 & 0.875 & 0.703 & 1.000 & 0.944 & 0.800 & 1.000 & 0.992 \\
F14  & H & 0.574 & 0.156 & 0.803 & 0.049 & 0.787 & 0.677 & 0.339 & 0.710 & 0.097 & 0.992 & 0.835 & 0.729 & 1.000 & 0.256 & 0.970 & 0.000 & 0.147 & 0.284 & 0.000 & 0.000 \\
F15  & H & 0.722 & 0.357 & 0.930 & 0.087 & 0.957 & 0.703 & 0.461 & 0.578 & 0.070 & 0.992 & 0.949 & 0.594 & 0.993 & 0.130 & 0.971 & 0.000 & 0.053 & 0.153 & 0.000 & 0.000 \\
F16  & N & 0.984 & 1.000 & 0.992 & 0.984 & 0.992 & 0.748 & 0.901 & 1.000 & 0.962 & 0.992 & 0.984 & 0.992 & 0.992 & 0.866 & 0.984 & 1.000 & 0.988 & 0.994 & 1.000 & 1.000 \\
F17  & N & 0.863 & 0.962 & 0.985 & 0.977 & 0.947 & 0.691 & 0.831 & 0.978 & 0.963 & 0.787 & 0.984 & 0.969 & 0.961 & 0.898 & 0.945 & 1.000 & 0.950 & 0.946 & 1.000 & 1.000 \\
F18  & N & 0.611 & 0.788 & 0.442 & 0.982 & 0.292 & 0.410 & 0.689 & 0.213 & 0.910 & 0.008 & 0.068 & 0.246 & 0.000 & 0.763 & 0.000 & 0.989 & 0.937 & 0.558 & 1.000 & 0.979 \\
F19  & N & 0.747 & 0.716 & 0.579 & 0.968 & 0.389 & 0.369 & 0.699 & 0.660 & 0.748 & 0.107 & 0.083 & 0.250 & 0.075 & 0.783 & 0.133 & 1.000 & 0.909 & 0.705 & 1.000 & 1.000 \\
F20  & N & 1.000 & 0.900 & 1.000 & 1.000 & 1.000 & 0.600 & 0.600 & 1.000 & 1.000 & 0.800 & 0.800 & 0.700 & 0.700 & 1.000 & 0.700 & 1.000 & 0.968 & 1.000 & 1.000 & 1.000 \\
F21  & N & 1.000 & 1.000 & 0.400 & 1.000 & 0.600 & 0.222 & 0.778 & 0.556 & 0.889 & 0.000 & 0.444 & 0.778 & 0.222 & 0.889 & 0.667 & 1.000 & 0.800 & 0.967 & 1.000 & 1.000 \\
F22  & N & 0.778 & 1.000 & 0.667 & 1.000 & 0.667 & 0.400 & 0.600 & 0.800 & 1.000 & 0.100 & 0.167 & 0.833 & 0.000 & 0.667 & 0.167 & 1.000 & 1.000 & 1.000 & 1.000 & 1.000 \\
F23  & H & 0.456 & 0.081 & 0.806 & 0.019 & 0.675 & -     & -     & -     & -     & -     & -     & -     & -     & -     & -     & -     & -     & -     & -     & -     \\
F24  & H & 0.351 & 0.145 & 0.634 & 0.015 & 0.664 & -     & -     & -     & -     & -     & -     & -     & -     & -     & -     & -     & -     & -     & -     & -     \\
F25  & H & 0.457 & 0.371 & 0.836 & 0.000 & 0.784 & -     & -     & -     & -     & -     & -     & -     & -     & -     & -     & -     & -     & -     & -     & -     \\
F26  & H & 0.539 & 0.023 & 0.711 & 0.023 & 0.758 & -     & -     & -     & -     & -     & -     & -     & -     & -     & -     & -     & -     & -     & -     & -     \\
F27  & H & 0.448 & 0.078 & 0.741 & 0.026 & 0.767 & -     & -     & -     & -     & -     & -     & -     & -     & -     & -     & -     & -     & -     & -     & -     \\
F32  & H & -     & -     & -     & -     & -     & -     & -     & -     & -     & -     & 0.571 & 0.217 & 0.683 & 0.180 & 0.621 & -     & -     & -     & -     & -     \\
F33  & H & -     & -     & -     & -     & -     & -     & -     & -     & -     & -     & 0.602 & 0.333 & 0.699 & 0.151 & 0.720 & -     & -     & -     & -     & -     \\
F34  & H & -     & -     & -     & -     & -     & -     & -     & -     & -     & -     & 0.589 & 0.411 & 0.795 & 0.041 & 0.740 & -     & -     & -     & -     & -     \\
\midrule
\multirow{3}{*}{Avg} & NH & 0.827 & 0.872 & 0.798 & 0.982 & 0.733 & 0.567 & 0.758 & 0.759 & 0.909 & 0.537 & 0.545 & 0.629 & 0.537 & 0.840 & 0.569 & 0.999 & 0.952 & 0.879 & 1.000 & 0.996 \\
                     & H & 0.501 & 0.206 & 0.801 & 0.023 & 0.747 & 0.583 & 0.397 & 0.547 & 0.079 & 0.905 & 0.794 & 0.418 & 0.929 & 0.173 & 0.866 & 0.000 & 0.089 & 0.198 & 0.000 & 0.005 \\
                     & both & 0.571 & 0.350 & 0.800 & 0.230 & 0.744 & 0.578 & 0.504 & 0.610 & 0.326 & 0.795 & 0.731 & 0.471 & 0.830 & 0.342 & 0.791 & 0.318 & 0.364 & 0.415 & 0.318 & 0.320 \\
\bottomrule
\end{tabular}
\caption{Accuracy of MultiSet fine-tuned LLMs when tested on \textsf{SGHateCheck} across functionalities functionalities(LL:LLaMA2, MB:mBert, MI:Mistral, SO:SEA-LION, SM: SeaLLM) and language. Gold Label (GL) of hate (H) and and non-hate (NH) are also shown. Please see Appendix ~\ref{app:annotation_functionality} for description of functionality number (F\#). Red numbers indicate an accuracy of less than 0.500, which is worse than chance.}
\label{tab:annotation_functionality}
\end{sidewaystable*}

Table~\ref{tab:annotation_functionality} shows the MultiSet fine-tuned LLMs' performance for various functionality tests. Upon closer examination of MultiSet fine-tuned models across various functional tests, it became evident that while all models demonstrated proficiency in identifying non-hateful content (\textbf{F16} and \textbf{F17}) and abuses targeting inanimate objects (\textbf{F20}), achieving accuracy scores over 0.600, disparities emerged in more nuanced categories. 

Despite their generally robust performance, Mistral and SeaLLM exhibited vulnerabilities in tests aimed at recognizing denunciations of hate speech (HS) (\textbf{F18}) that included quotations of the original HS, where their accuracy dropped to 0.219 or lower. This issue was more pronounced in Mandarin, where the models sometimes completely failed to detect such nuances, as evidenced by a zero accuracy score. Additionally, these models performed poorly in tests focusing on abuse directed at non-target individuals and groups (\textbf{F21} and \textbf{F22}), with their accuracy falling below 0.667.



Excluding results for Tamil, where all models uniformly underperformed, the data revealed a lack of consistency in model performance across languages within identical functional groups. This inconsistency did not follow a discernible pattern related to the language of the test cases. For example, SeaLLM's performance varied across languages; it fared better in Malay compared to Singlish and Mandarin. However, its weakest functional categories in Malay were significantly outperformed in other languages, underscoring the complex interplay between model training, linguistic context, and the inherent challenges of accurately classifying nuanced HS across diverse languages.




\subsection{How do the fine-tuned models perform across target groups?}
\definecolor{Orange}{RGB}{255,158,62}
\definecolor{yellow}{rgb}{1.0, 0.88, 0.21}
\definecolor{yellowgreen}{rgb}{0.79, 0.86, 0.54}
\definecolor{green}{rgb}{0.2, 0.8, 0.2}
\definecolor{red}{rgb}{0.84, 0.23, 0.24}

\begin{table*}[ht]
\small
    \centering
    {
    \begin{tabular}{ll*{5}{c}|c}
\toprule
Prt. Grp.               &    Target    & LLaMA2 & mBert & Mistral & SEA-LION & SeaLLM & Average\\
\hline
Age              & Seniors             & 0.430 & 0.340 & 0.462 & 0.256 & 0.456 & 0.389 \\
\hline
\multirow{2}{*}{Disability}       & Mentally Ill        & 0.426 & 0.332 & 0.578 & 0.239 & 0.586 & 0.432 \\
       & Physically Disabled & 0.478 & 0.410 & 0.534 & 0.243 & 0.563 & 0.446 \\
\hline
\multirow{3}{*}{Gender/Sexuality}  & Homosexual          & 0.538 & 0.384 & 0.648 & 0.246 & 0.609 & 0.485 \\
 & Transsexual          & 0.478 & 0.376 & 0.614 & 0.258 & 0.598 & 0.465 \\
 & Women               & 0.553 & 0.474 & 0.648 & 0.246 & 0.623 & 0.509 \\
\hline
Nationality      & Immigrants          & 0.490 & 0.357 & 0.658 & 0.245 & 0.592 & 0.469 \\
\hline
\multirow{3}{*}{Race}             & Chinese             & 0.545 & 0.429 & 0.699 & 0.264 & 0.634 & 0.514 \\
             & Indians             & 0.516 & 0.369 & 0.651 & 0.258 & 0.622 & 0.483 \\
             & Malay               & 0.523 & 0.425 & 0.639 & 0.268 & 0.630 & 0.497 \\
\hline
\multirow{4}{*}{Religion}         & Buddhist            & 0.437 & 0.376 & 0.544 & 0.271 & 0.576 & 0.441 \\
         & Christian           & 0.464 & 0.347 & 0.596 & 0.260 & 0.603 & 0.454 \\
         & Hindu               & 0.461 & 0.355 & 0.608 & 0.289 & 0.573 & 0.457 \\
         & Muslim              & 0.564 & 0.487 & 0.720 & 0.253 & 0.636 & 0.532 \\
\hline 
& Average & 0.493 & 0.390 & 0.614 & 0.257 & 0.593 \\ 
\bottomrule
\end{tabular}
    \caption{F1 scores for protected groups (Prt. Grp.) and its target placeholders in Singlish, Mandarin, Malay and Tamil for MultiSet fine-tuned models}
    \label{tab:annotation_protected_groups}
    }
\end{table*}

Table~\ref{tab:annotation_protected_groups} shows the MultiSet fine-tuned LLMs' performance on \textsf{SGHateCheck} breakdown by protected groups. The more effective LLMs, specifically Mistral and SeaLLM, showcased superior performance with an average F1 score exceeding 0.593. In contrast, mBert and SEA-LION lagged significantly, with their scores not surpassing 0.390. Analyzing performance across different target groups, it was observed that representations of seniors received the lowest average F1 score of 0.389. Conversely, categories pertaining to the Muslims were identified with the highest scoring, reaching up to 0.532. Notably, among racial groups, Indians and, within religious categories, Buddhists were the lowest scoring targets, indicating potential areas for model improvement.






\subsection{How does the performance on SGHateCheck compare with that on HateCheck and MHC?}

To evaluate \textsf{SGHateCheck}'s efficacy against non-localized counterparts, we tested models trained with MultiSet on HateCheck and MHC's Mandarin dataset (results shown in Table~\ref{tab:MHC_SHCS}. Initial comparisons on language pairs (\textsf{SGHateCheck} Mandarin vs. MHC Mandarin, and \textsf{SGHateCheck} Singlish vs. HateCheck) show similar average macro F1 scores. However, a deeper analysis into specific functionalities reveals significant differences. For instance, performance on \textsf{SGHateCheck} Mandarin showed notable discrepancies in certain areas compared to MHC Mandarin, and similarly, \textsf{SGHateCheck} Singlish diverged significantly from HateCheck in classes related to non-hateful group identifiers, highlighting the unique challenges and contributions of \textsf{SGHateCheck} in detecting HS within localized contexts.



\begin{table}[]
\small
\begin{tabular}{>{\arraybackslash}m{1cm}*{4}{>{\centering\arraybackslash}m{1cm}}}
\toprule
F\#     & MHCM & SHCM & HC & SHCS \\ \hline
F7        & 0.224        & 0.421                & 0.270     & 0.208    \\
F16       & 0.690        & 0.490                & 0.799     & 0.598    \\
F17       & 0.481        & 0.487                & 0.799     & 0.486    \\
F19       & 0.308        & 0.180                & 0.475     & 0.397    \\ \hline
Overall F1 & 0.564        & 0.585                & 0.535     & 0.507    \\ \bottomrule

\end{tabular}

\caption{F1 scores of selected functionalities (F\#) for MHC Mandarin (MHCM), \textsf{SGHateCheck} Mandarin (SHCM), HateCheck (HC) and \textsf{SGHateCheck} Singlish (SHCS). Please see Appendix ~\ref{app:annotation_functionality} for description of functionality number (F\#)}
\label{tab:MHC_SHCS}
\end{table}

\subsection{Discussion}

The nuanced findings from our experiments with \textsf{SGHateCheck} offer valuable insights into the landscape of HS detection models. 
Overall, models perform better with straightforward, direct representations of hateful speech (HS) and non-hateful test cases, but struggle in more complex scenarios, such as when HS is employed illustratively in denunciations. This observation aligns with our hypothesis that the limitations identified in HateCheck and MHC are also present in the Singapore context.


Comparing the different models we tested, Mistral 7B's standout performance raises intriguing questions, especially given its efficiency across diverse languages and tasks, save for a couple of specific functionalities in Mandarin. This exception not only piques interest but also marks an area ripe for in-depth analysis to uncover underlying reasons behind this deviation.

The observed bias towards non-hateful classifications in models like mBert and SEA-LION, despite mBert's strong performance in isolated tests, brings to light the critical role of \textsf{SGHateCheck} in identifying and mitigating model biases. This discrepancy highlights the tool's effectiveness in revealing blind spots that traditional held-out tests might overlook, emphasizing the importance of comprehensive testing beyond standard datasets.

Moreover, the benefits of a varied fine-tuning dataset become evident, aligning with the theory that cross-lingual transfer can enhance model performance. However, this improvement isn't uniformly observed across all languages, particularly in Tamil, where the expected boost in model effectiveness was minimal. Such variability underscores the complexity of language-specific biases and the challenges in generalizing model improvements across diverse linguistic contexts.

Finally, the comparative analysis between \textsf{SGHateCheck} and benchmarks like MHC Mandarin and HateCheck uncovers specific functional areas where models underperform, despite seemingly similar overall effectiveness. This discrepancy underscores the necessity for targeted functional tests to precisely diagnose and address model weaknesses, reinforcing the importance of localization and context-specificity in developing robust HS detection systems.

\section{Related Work}

\subsection{English Hate Speech Datasets}

Hate speech (HS) includes expressions that attack or demean groups based on characteristics such as race, religion, ethnic origin, sexual orientation, disability, or gender. 
Researchers have developed numerous datasets to study HS across different platforms, with a focus on explicit text-based \cite{pamungkas2020you,founta2018large,waseem-hovy-2016-hateful,davidson2017automated}, implicit text-based \cite{mathew2021hatexplain,elsherief2021latent}, and multimodal hate speech \cite{kiela2020fhm,fersini2022mami,hee2023decoding}. Recent efforts have also involved the development of generative methods to create adversarial datasets for improved HS detection. However, ensuring the quality and consistency of annotations in naturally collected data poses a significant challenge \cite{awal2020analyzing}. Recent studies have delved into diagnostic methods that provide robust functional tests to systematically evaluate hate speech detection models \cite{rottger-etal-2021-hatecheck,rottger-etal-2022-multilingual}.

\subsection{Non-English Hate Speech Datasets} Given the scarcity of datasets in non-English languages, there have been attempts to do zero-shot cross-lingual HS detection but model performance has been found to be lacking \cite{pelicon2021investigating,nozza-2021-exposing,bigoulaeva-etal-2021-cross}. Therefore to bridge this gap, we see several datasets curated for specific regions \cite{moon2020beep,chinese_training_dataset}. 

 There has been recent interest in application of hateful content moderation in the SEA region, involving some of the low resource languages. This has led to several new datasets created for this purpose, notably Indonesian hate speech datasets (\citet{related_indo_review,related_indo_1,related_indo_2}), Thai Dataset (\citet{related_thai}) and Vietnamese HS dataset (\citet{related_viet_1}). The data is collected from social media such as twitter and human annotator provide binary hateful/non-hate labels. With SGHateCheck, we extend the idea of diagnostic dataset of HateCheck to SEA region.

 \subsection{Hate Speech Detection Models}

 Hate speech (HS) detection has been a significant area of research, leveraging natural language processing (NLP) techniques. Existing studies have developed NLP methods using deep learning to train models for detecting hate speech, which includes learning multi-faceted text representations \cite{cao2020deephate,mahmud2023deep} and fine-tuning transformer-based models \cite{awal2021angrybert, caselli2021hatebert}. Additionally, researchers have explored other approaches such as using model-agnostic meta-learning for detecting hate speech across multiple languages \cite{awal2023model}, and analyzing network propagation and conversation threads to identify instances of hate speech \cite{lin2021early,meng2023predicting}. Furthermore, with the recent emergence of large language models (LLMs), there is increasing exploration into using these LLMs for detecting and explaining hate speech \cite{han2023evaluating}. Consequently, there is a growing need to systematically evaluate the robustness of these hate speech detection systems.

\section{Conclusion}

The unveiling of \textsf{SGHateCheck} marks a pivotal advancement in HS detection research, bridging the gap between global methodologies and Singapore's distinct sociolinguistic landscape. By integrating Singlish, Malay, Tamil, and a culturally adapted Mandarin dataset, \textsf{SGHateCheck} extends beyond the foundational frameworks provided by HateCheck and MHC. This expansion results in a comprehensive suite of over 21,152 test cases, with 11,373 meticulously annotated, encompassing both hateful and non-hateful content. This breadth and depth offer a nuanced platform for evaluating HS detection models, enabling a detailed analysis of their capabilities and limitations across a spectrum of linguistic and cultural contexts.

\textsf{SGHateCheck} serves as a diagnostic tool, rigorously testing five models fine-tuned on diverse HS datasets in English, Mandarin, and Malay. The findings reveal a significant bias in models towards classifying ambiguous cases as non-hateful, particularly in languages or dialects not included in their training data. This limitation underscores the importance of comprehensive and localized testing frameworks like \textsf{SGHateCheck}, which can uncover biases that conventional held-out tests may overlook.

Amidst a research landscape traditionally dominated by Western socio-linguistic norms, \textsf{SGHateCheck} pioneers a shift towards more localized interpretations of HS. This shift is crucial for the development of detection models that are both effective and sensitive to the nuances of regional languages and dialects, especially in the linguistically diverse Southeast Asian region. Through \textsf{SGHateCheck}, we aspire to inspire and catalyze further research into HS detection in low-resource languages, fostering a more inclusive and equitable digital discourse.




\section{Limitation}

Building on HateCheck and MHC, \textsf{SGHateCheck} adapts their framework to Singapore's unique context but also inherits some limitations, such as focusing more on model weaknesses rather than strengths and not accounting for external context or the full spectrum of protected groups. 
The use of fixed template-placeholder pairs to generate test cases significantly restricts their flexibility. As a result, they fail to effectively represent certain specific forms of hate, such as demeaning a transgender individual.
The linguistic diversity and code-switching prevalent in Singapore pose additional challenges, making the monolingual approach less reflective of real-world hate speech usage. Moreover, the direct translation of templates without local nuances may not fully capture the local expression of hate, highlighting the need for a more nuanced approach to truly reflect Singapore's sociolinguistic landscape.





\section*{Acknowledgments}
This research/project is supported by Ministry of Education, Singapore, under its Academic Research Fund (AcRF) Tier 2. Any opinions, findings and conclusions or recommendations expressed in this material are those of the authors and do not reflect the views of the Ministry of Education, Singapore.


\bibliography{custom}

\newpage
\appendix



\section{Data Statement}
\subsection{Curation Rationale}

\textsf{SGHateCheck} functional test dataset made specially to test for the sociolinguistical context of Singapore. Templates from MHC and HateCheck were translated by language experts with the help of machine generated cases. In total, 21,152 test-cases were generated and 11,373 test cases were annotated as hateful, non-hateful or nonsensical. 

\subsection{Language Variety}

\textsf{SGHateCheck} covers Singlish, Malay, Mandarin and Tamil.

\subsection{Translator and Annotators Proficiency and Demographics}
\label{appendix:helper}

All translators and annotators have the target language proficiency (Studied as a subject in school for at least 10 years and/or use it in a family setting) and use them in social situations (Read and/or write it in social media and/or use it with family and/or friends). 

Before participating, all annotators were briefed about the definition of HS and protected groups in the study. We screened them on a hateful/non-hate classification task on a sample dataset, for the respective languages.

All translators and annotators are fluent in English in addition to the target language. They were in their 20s and were studying for their Bachelors or Masters. 5 of the 8 translators and 8 of the 18 annotators are females. 

\subsection{Data Creation Period}

Translations were done between November 2023 and February 2024. Annotations were created between January 2024 and March 2024. 

\subsection{Functionality and Annotation}
\label{app:annotation_functionality}

Table ~\ref{app:annotation_functionality} shows the full description of each functionality, as well as the number of annotations in each of them.

\section{Inter Annotator Agreement and Test Case Exclusion}
\label{appendix:iaa}

To ensure the quality of the test cases used in the experiments, we excluded ambiguous test cases and calculated the inter-annotator agreement (IAA) for the remaining test cases.

\subsection{Inter-Annotator Agreement}


The IAA score for each language is calculated using Krippendorff's $\alpha$ \cite{krippendorff2018content}, as shown Table \ref{tab:iaa_scores}. All languages have an IAA score greater than 0.667, indicating an acceptable level of agreement.

\subsection{Excluded Test Cases}


Firstly, we treat test cases lacking majority consensus as ambiguous and exclude them from our experiments (``Undetermined''). Singlish, Malay, and Mandarin each have fewer than ten cases of this nature. Conversely, Tamil, which has only two annotations per test case, exhibits a significantly higher number of these ambiguous cases.


Secondly, if the labels of test cases do not match those of their corresponding templates, the test cases are deemed ambiguous and are excluded from the experiments (``Mismatch''). All languages have less than 100 instances of such cases.


The overview of annotated test cases, unanimous annotations, undecided annotations and annotations that do not match 'Gold Labels' can be found in Table \ref{tab:annotation_metrics}.

\begin{table*}[t!]
\small
\centering
\begin{tabular}{r*{6}{c}}
\toprule
& \multicolumn{4}{c}{Compilation} & \multicolumn{2}{c}{Annotations} \\
\cmidrule(lr){2-5} \cmidrule(lr){6-7}
Lang.    & Unanimous. & 2 out of 3 & Undetermined & Mismatch & Retained & Excluded  \\
\midrule
Singlish & 2695 & 276    & 3    & 38  & 2933 & 41 \\
Malay    & 2041 & 207    & 5    & 40  & 2208 & 45     \\
Mandarin & 2330 & 511    & 7    & 64  & 2777 & 71      \\
Tamil    & 2559 & -      & 292  & 83  & 2476 & 375  \\
\bottomrule
\end{tabular}
\caption{Breakdown of annotation compilation. \textit{Unanimous} indicates that all annotators agreed on the same annotation. \textit{2 out of 3} means two out of three annotators agreed (N/A for Tamil because each test cases only had 2 annotations). \textit{Undetermined} denotes cases where each annotators disagree completely and chose different options. \textit{Mismatch} occurs when the labels of test cases differ from those of their corresponding templates. \textit{Retained} represents the number of test cases validated as robust and used in the experiments, while \textit{Rejected} denotes those excluded due to ambiguity. }
\label{tab:annotation_metrics}
\end{table*}

\begin{table}[t!]
\small
\centering
\begin{tabular}{rcc}
\toprule
Language    & Krippendorff's $\alpha$ \\
\midrule
Singlish & 0.800 \\
Malay    & 0.817 \\
Mandarin & 0.682 \\
Tamil    & 0.672 \\
\bottomrule
\end{tabular}
\caption{The inter-annotator agreement scores for individual languages.}
\label{tab:iaa_scores}
\end{table}

\section{Finetuning Details}
\label{appendix:finetuning}

For all models, the hardware used are NVIDIA GeForce RTX 3090  with 24gb of memory. 

\subsection{\citet{waseem-hovy-2016-hateful}}
Labelled English HS dataset used in EngSet and MultiSet fine-tuning.
\subsubsection{Sampling} First, a manual search of common slur words was used to obtain a basket of frequently occurring terms. Next, terms were fed into the Twitter search API to collect the data. In total 136,052 tweets were collected and 16,914 tweets were annotated. 
\subsubsection{Annotation} The annotations were done by the authors and reviewed by a 25 year old female gender studies student. The tweets were labelled one of All, Racism, Sexism and Neither. The inter-annotator agreement had a Cohen's $\kappa$ of 0.84. 
\subsubsection{Data Used} 16,038 of 16,914 tweets were used (31.1\% of tweets used are hateful). Some tweets became inaccessible at the time of data collation. 
\subsubsection{Definition of HS} A list of 11 HS identifiers were identified by the authors. The criteria are partially derived by negating the privileges observed in \citet{2003-06587-018}, where they occur as ways to highlight importance, ensure an audience, and ensure safety for white people, and partially derived from applying common sense.

\subsection{\citet{mathew2021hatexplain}}
Labelled English HS dataset used in EngSet and MultiSet fine-tuning
\subsubsection{Sampling} Dataset was sourced from Twitter \cite{hateexplain_twitter_source, hateexplain_twitter_gab_source, hateexplain_twitter_source_2} and Gab \cite{hateexplain_twitter_gab_source}. The twitter dataset consists of 1\% of randomly collected tweets from January 2019 to June 2020. Reposts and duplicates were removed, and usernames were masked. In total, 9,055 entries were taken from twitter and 11,093 were taken from Gab. 
\subsubsection{Annotation} MTurks with high HIT Approval Rate and HIT Approved were used for annotation. Each entry was annotated 3 times, and labelled Hateful (29.5\% of the dataset), Offensive, Normal or Undecided. The Krippendorff's $\alpha$ was 0.46. 
\subsubsection{Data Used} 15.4k annotations in the training data split used. Of the 4 possible labels used, cases with the 'Hateful' label were labelled as hateful, the rest were considered non-hateful.
\subsubsection{Definition of HS} The definition is taken from \citet{hateexplain_twitter_source} which is \textit{language that is used to expresses hatred towards a targeted group or is intended to be derogatory, to humiliate, or to insult the members of the group}. The target groups used in HateXplain are Race, Religion, Gender, Sexual Orientation and Miscellaneous. 

\subsection{\citet{malay_dataset}}
Labelled Malay HS dataset used in MultiSet fine-tuning
\subsubsection{Sampling} Data was gathered using the Twitter streaming API and Search API using a basket of keywords commonly associated with cyberbullying \cite{malay_dataset_words}. The texts were removed if it is a retweet, is not written in Malay, has a URL or has less than 10 characters. 
\subsubsection{Annotation} An initial group of annotators annotated 300 tweets. These tweets were used to train and select 3 annotators fluent in Malay as main annotators. Where the annotators could not come up with a majority decision, a third annotator was involved. The inter-annotator agreement had a Fleiss' $\kappa$ of 0.85. 4,892 tweets were annotated as one of non-hateful or hateful (38.6\%). 
\subsubsection{Data Used} All 4,892 samples were used for training 
\subsubsection{Definition of HS} The definition is taken from \citet{United_Nations_hates_speech} which is \textit{any kind of communication in speech, writing or behaviour, that attacks or uses prerogative or discriminatory language with reference to a person or a group on the basis of who they are}. The target groups identified are taken from Twitter: \textit{race, ethnicity, national origin, caste, sexual orientation, gender, gender identity, religious affiliation, age, disability, or serious disease}. 

\subsection{\citet{chinese_training_dataset}}
Labelled Mandarin dataset used in MultiSet fine-tuning
\subsubsection{Sampling} Keywords and descriptive words of target groups were identified and crawled in widely discussed sub-topics in Zhihu, a social media site from China.
\subsubsection{Annotation} 17 native Mandarin speakers (9 males, 8 femailes) from various regions in China were used to annotate the data as Offensive or Non-Offensive. The data was first pre-annotated by machine, and corrected by annotators. Each case was processed by 3 annotators, and the majority annotation becomes the final annotation. The inter-annotator agreement had a Fleiss' $\kappa$ of 2-class is 0.819. 
\subsubsection{Data Used} A Model-in-the-loop system was used to split data into Train, Dev and Train. 25,726 (50\% offensive) cases in the Train dataset was used in training.
\subsubsection{Definition of HS} As defined in the questionaire given to annotators, \textit{Does the text contains any form of targeted offense to individuals or groups? It includes veiled or direct offensive content expressing rudeness, disrespect, insults, threats and profanity based on something such as race, religion, sex or sexual orientation}. 

\section{Prompts and Result Processing}
\subsection{LLM Prompts}
\label{app:benchmark_prompts}
The following prompt was used for training of the models

\begin{lstlisting}
"""Label the given Input as 0 or 1, where 0 is not hateful and 1 is hateful.
### Input:
{Test Case}
### Label:
{Label}"""
\end{lstlisting}

Where \{Case\} is a case from the dataset, and \{Label\} is 0 (representing non-hate) or 1 (representing hate).

The following prompt was used for evaluation of models

\begin{lstlisting}
"""Label the given Input as 0 or 1, where 0 is not hateful and 1 is hateful.
### Input:
{Test Case}
### Label:
"""
\end{lstlisting}

Where \{Case\} is a case from the dataset, and \{Label\} is 0 (representing non-hate) or 1 (representing hate).

For post processing, the last character of the prediction is read 0 (representing non-hate) or 1 (representing hate).

\subsection{LLM Rejected Ouputs}
\label{app:benchmark_reject}

There are cases where output was a repetition of the prompt without any further prediction, or an empty prompts. Models with bad outputs, together with the number of occurrence from the corresponding test sets are as follows 

LLaMA2 trained with EngSet:
\begin{itemize}
    \item 24 from \textsf{SGHateCheck} Tamil
    \item 2 from \textsf{SGHateCheck} Mandarin
\end{itemize}

SEA-LION trained with EngSet
\begin{itemize}
    \item 3 from HateCheck
    \item 1723 from MHC Mandarin
    \item 15 from \textsf{SGHateCheck} Singlish
    \item 309 from \textsf{SGHateCheck} Malay
    \item 1979 from \textsf{SGHateCheck} Tamil
    \item 1693 from \textsf{SGHateCheck} Mandarin
\end{itemize}

SEA-LION trained with MultiSet: 
\begin{itemize}
    \item 1 from \textsf{SGHateCheck} Tamil
\end{itemize}

\begin{table*}[t!]
    \begin{tabular}{m{2cm}|m{6cm}c|cccc}
       \hline 
        \multirow{2}{2cm}{\textbf{Func. Class}} & \textbf{Functionality} & \multirow{2}{2cm}{\centering\textbf{Gold Label}} & \multicolumn{4}{c}{\textbf{\# of Annotated Cases}} \\
        & & & SS & MS & ZH & TA \\
        \hline
        \multirow{4}{*}{Derogation}  & \textbf{F1}: Expression of strong negative emotions
        (explicit) & hateful & 140	&	126	&	140	&	140
        \\
         & \textbf{F2}: Description using very negative attributes (explicit) & hateful & 84	&	112	&	112	&	210
        \\
         & \textbf{F3}: Dehumanisation (explicit)
        (explicit) & hateful & 131	&	132	&	126	&	146
        \\
         & \textbf{F4}: Implicit derogation & hateful & 303	&	140	&	139	&	140
        \\
        \hline

        \multirow{2}{2cm}{Threat. language}  & \textbf{F5}: Direct threat
        (explicit) & hateful & 131	&	119	&	140	&	140
        \\
        & \textbf{F6}: Threat as normative statement & hateful & 140	&	140	&	140	&	168 
        \\
        \hline
        Slurs & \textbf{F7}: Hate expressed using slur & hateful & 12	&	20	&	16	&	18
        \\
        \hline

        \multirow{2}{2cm}{Profanity usage} & \textbf{F8}: Hate expressed using profanity & hateful & 140	&	140	&	140	&	118
        \\
         & \textbf{F9}: Non-hateful use of profanity & non-hate & 10	&	10	&	10	&	46
        \\
        \hline

        Pronoun reference & \textbf{F10}: Hate expressed through reference in
        subsequent clauses & hateful & 140	&	140	&	140	&	126
        \\
         & \textbf{F11}: Hate expressed through reference in subsequent sentences & non-hate & 140	&	140	&	140	&	196
        \\
        \hline

        Negation & \textbf{F12}: Hate expressed using negated positive statement & hateful & 113	&	116	&	140	&	152
        \\
         & \textbf{F13}: Non-hate expressed using negated hateful statement
        & non-hate & 131	&	132	&	140	&	168
        \\
        \hline

        \multirow{2}{2cm}{Phrasing} & \textbf{F14}: Hate phrased as a question & hateful & 122	&	124	&	140	&	157
        \\
         & \textbf{F15}: Hate phrased as an opinion & hateful & 117	&	132	&	140	&	160
        \\
        \hline

        \multirow{2}{2cm}{Non-hateful group identifier} & \textbf{F16}: Neutral statements using protected group identifiers & non-hate & 131	&	132	&	140	&	171
        \\
         & \textbf{F17}: Positive statements using protected group identifiers & non-hate & 140	&	140	&	140	&	269
        \\
        \hline

         Counter speech & \textbf{F18}: Denouncements of hate that quote it & non-hate & 118	&	122	&	120	&	118 \\
         & \textbf{F19}: Denouncements of hate that make direct
        reference to it & non-hate & 100	&	106	&	362	&	82 
        \\
        \hline

        \multirow{2}{2cm}{Abuse against non-protected targets} & \textbf{F20}: Abuse targeted at objects & non-hate & 10	&	10	&	10	&	37
        \\
         & \textbf{F21}: Abuse targeted at individuals (not as member of a protected group) & non-hate & 10	&	10	&	10	&	36
        \\
        & \textbf{F22}: Abuse targeted at non-protected groups (e.g. professions) & non-hate & 10	&	10	&	10	&	42
        \\
        \hline

        \multirow{2}{2cm}{Spelling variations} & \textbf{F23}: Swaps of adjacent characters & hateful & 150	&	-	&	-	&	-
        \\
         & \textbf{F24}: Missing characters & hateful & 131	&	-	&	-	&	-
        \\
        & \textbf{F25}: Missing word boundaries & hateful & 118	&	-	&	-	&	-
        \\
        & \textbf{F26}: Added spaces between chars & hateful & 115	&	-	&	-	&	-
        \\
        & \textbf{F27}: Leet speak spellings & hateful & 87	&	-	&	-	&	-
        \\
        & \textbf{F32}: ZH: Homophone char. replacement & hateful & -	&	-	&	140	&	-
        \\
        & \textbf{F33}: ZH: Character decomposition & hateful & -	&	-	&	58	&	-
        \\
        & \textbf{F34}: ZH: Pinyin spelling & hateful & -	&	-	&	55	&	-
        \\
        \hline
         & \textbf{Total} & non-hate & 618	&	656	&	656	&	865
        \\
         &  & hate & 2298	&	1552	&	2083	&	1724
        \\
        &  & Total & 2974	&	2253	&	2848	&	2851
        \\
        \hline
    \end{tabular}
    \label{tab:annotation_functionality_count}
    \caption{ Number of test-cases annotated in \textsf{SGHateCheck} across functionalities. Also shown in this table is the functional class which the functionalities belong to, its functionality number and gold labels.}
\end{table*}


\end{document}